# Comparative Analysis of Machine Learning Models for Predicting Travel Time

**Aboah Armstrong**
*Department of Civil and Environmental Engineering- University of Missouri, Columbia*
*e-mail: aa5mv@umsystem.edu*

**Arthur Elizabeth**
*Department of Civil and Environmental Engineering- NDSU*
*e-mail: Elizabeth.arthur@ndsu.edu*

**Abstract**
In this paper, five different deep learning models are being compared for predicting travel time. These models are autoregressive integrated moving average (ARIMA) model, recurrent neural network (RNN) model, autoregressive (AR) model, Long-short term memory (LSTM) model, and gated recurrent units (GRU) model. The aim of this study is to investigate the performance of each developed model for forecasting travel time. The dataset used in this paper consists of travel time and travel speed information from the state of Missouri. The learning rate used for building each model was varied from 0.0001-0.01. The best learning rate was found to be 0.001. The study concluded that the ARIMA model was the best model architecture for travel time prediction and forecasting.

## 1. Introduction

Intelligent Transportation System (ITS) is a fast-growing niche of research which has gain popularity in recent times due to the overall advancement in technology. One of its integral components is travel time estimation and forecasting. Travel time forecasting helps in traffic control by means of reducing traffic congestion and providing real-time estimation of traffic estimation. Several studies have adopted different methodologies in predicting or forecasting travel time. Despite, the numerous studies conducted, there is no concrete conclusion on the approach which produces the best travel time estimation with minimum errors. Therefore, the authors in the study seek to compare five different models often used for predicting travel time on the same dataset. These models are autoregressive integrated moving average (ARIMA) model, recurrent neural network (RNN) model, autoregressive (AR) model, Long-short term memory (LSTM) model, and gated recurrent units (GRU) model.

The aim of this project is to investigate the performance of each developed model for forecasting travel time. The data used for the study was collected by the State of Missouri on Interstate-70 (I-70). To evaluate the effectiveness of the developed models, the study compared all five models using root-mean-squared-error (RMSE) and mean-absolute-error (MAE).

The rest of this paper is organized as follows. Section two presents a review of previous studies. The dataset used in this study is presented in Section three while Section four presents discussion and analysis of results gotten from the developed models. The conclusions and findings of the study are finally presented in Section five.

## 2. Related work

Numerous studies have been conducted so far as estimation of travel time is of concern. Different models have been built for travel time prediction and forecasting using different modeling techniques. Some of these techniques include parametric and non-parametric approaches, machine learning methods, and deep learning methods. This section reviews some of the specific techniques and models that have been developed for travel time prediction and forecasting as show in Table 1.

**Table 1: Summary of related works**

| Reference | Year | Approached used | Prediction Parameter | Major findings |
|---|---|---|---|---|
| Ide et al. [1] | 2009 | Probabilistic approach | Travel Time | Findings:<br>1. The study introduced the use of kernel string to represent the similarity between paths.<br>2. The use of Gaussian procession for travel time prediction |
| Wu et al. [2] | 2003 | Support Vector Regression (SVR) | Travel Time | Findings:<br>1. The studies showed that SVR produced better results compared to Artificial Neural Networks |
| Billings et al. [3] | 2006 | autoregressive integrated moving average (ARIMA) | Travel Time | Findings:<br>1. LSTM had the minimum value of MRE compared to the other models. |
| HU et al. [4] | 2016 | non-linear time-series modeling, linear model, support vector regression (SVR) model | Travel Time | Findings:<br>1. In this study, SVR performed better than multiple linear regression model and historical average models. |
| Duan et al. [5] | 2016 | Long Short-Term Memory neural network (LSTM) | Travel Time | Findings:<br>1. The study found out that deep learning models with relational sequence are promising in traffic time prediction |
| Liu et al. [6] | 2017 | LSTM, linear regression, Rigde-regression, Lasso Regression, deep neural network (DNN) and ARIMA | Travel Time | Findings:<br>1. LSTM-DNN model performed better than all other models. |

## 3. Data

The dataset used in this project consists of travel time and travel speed information for the state of Missouri. The dataset is made up of 48,605,476 rows and 7 columns. There are 542 unique Traffic Message Channel (TMC) code of which 10 unique TMC codes were used in this project. The project focused on predicting and forecasting travel time. Figure 1 shows a plot of the time frequency domain.

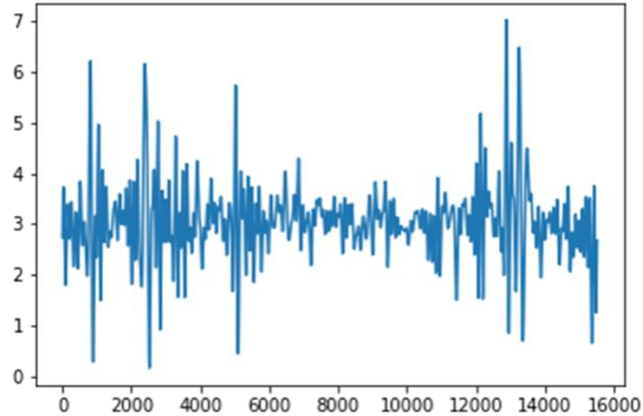

Figure 1: A plot of time frequency

*3.1 Data structure:*
The data was divided into training and validation sets. Each dataset was shaped in the format N x T x D. Where N is the size of the data for training or validation, T is the number of past information to be used for the prediction, and D is the number of different parameters to be used to predict the dependent variable. Since only one variable was considered, the number of input parameters was 1 for all the models and the number of outputs was also 1. Table 2 summarizes the shape of datasets (X_Shape and Y_shape), the number of input (n_inputs) the number of hidden layers used (n_hidden), the number of RNN layers used (n_rnnlayers) and the number of outputs expected (n_outputs).

Table 2: Data structure and model architecture

| Model | X_shape | Y_shape | n_inputs | n_hidden | n_rnnlayers | n_outputs |
|---|---|---|---|---|---|---|
| LSTM | 15502, 10, 1 | 15502, 1 | 1 | 25 | 2 | 1 |
| RNN | 15502, 10, 1 | 15502, 1 | 1 | 25 | 2 | 1 |
| ARIMA – AR (10), MA (5) | 15502, 15 | 15502, 1 | 1 | x | x | 1 |
| AR | 15502, 10 | 15502, 1 | 1 | x | x | 1 |
| GRU | 15502, 10, 1 | 15502, 1 | 1 | 25 | 2 | 1 |

## 4. Results and discussion

*4.1 Training and Validation losses:*
All models were trained for 5000 epochs. The learning rate used for building each model was varied from 0.0001-0.01. The best learning rate was found to be 0.001. From Figure 2, all training and testing losses converged.

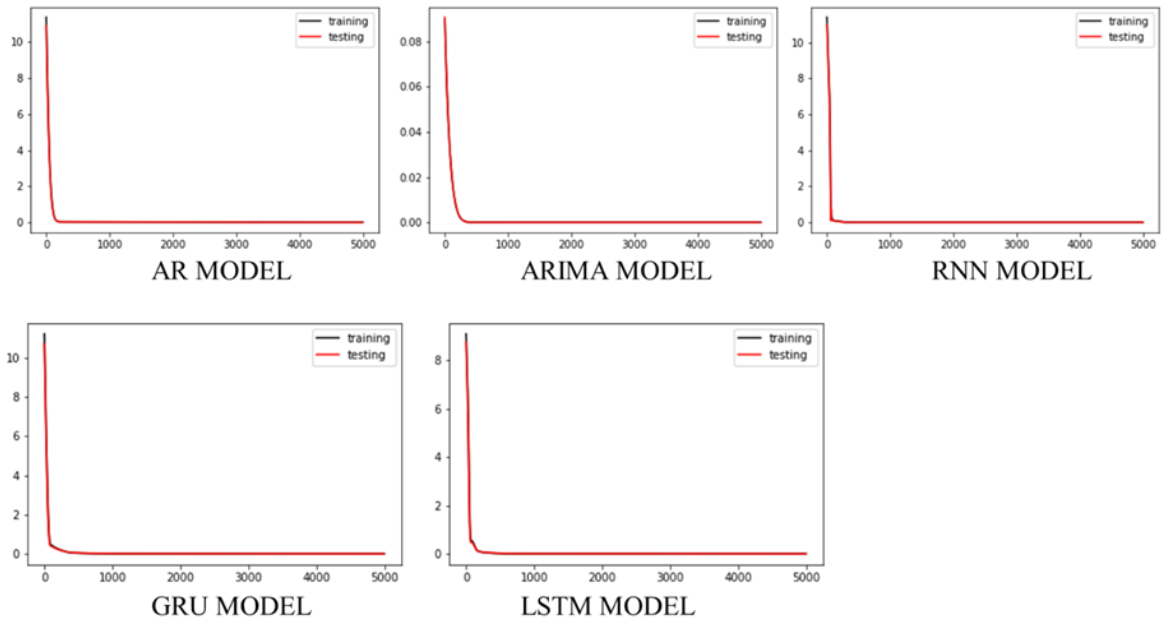

Figure 2: Plot of train and test losses for the various models

### 4.2 Evaluation of developed models using Mean Absolute Error (MAE) and Root Mean Squared Error (RMSE):

The developed models were evaluated using MAE and RMSE. The autoregressive integrated moving average (ARIMA) model gave the least MAE value while GRU gave the greatest MAE value. The RNN model performed relatively better than LSTM model and the AR model. When the developed models were evaluated using the RMSE, the ARIMA had the least RMSE while the RNN model had the highest value of RMSE. The AR model was relatively giving a better result in terms of RMSE compared to the LSTM model and the GRU model. Figure 3 and Figure 4 show a bar chart comparing the various MAE and RMSE of the developed models.

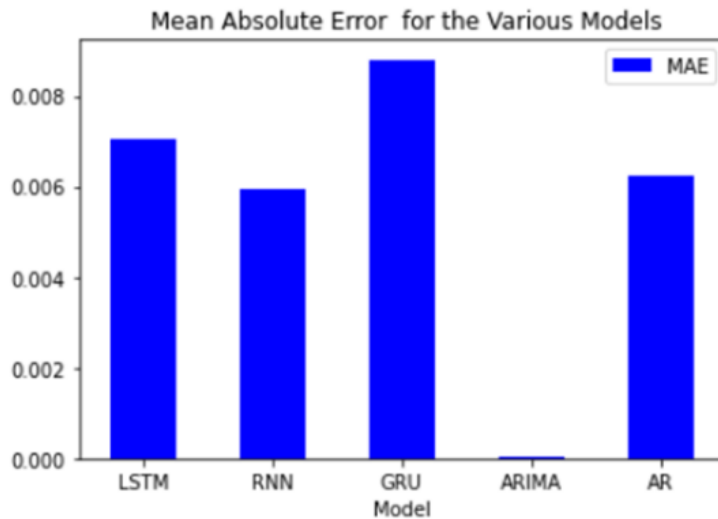

Figure 3: Evaluating the performance of the developed models use MAE

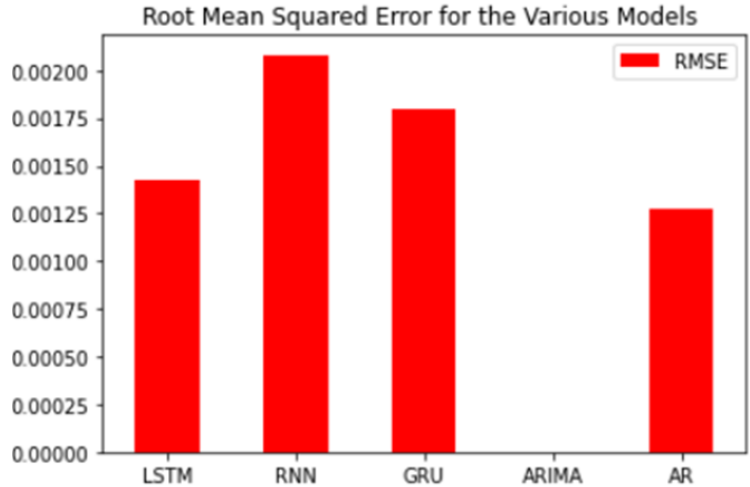

Figure 4: Evaluating the performance of the developed models use RMSE.

### *4.3 Comparing One-step forecasting for all developed models:*

Figure 5 shows the various graphs for one-step forecasting for the developed models. The effectiveness of each model was evaluated using MAE and RMSE. The AR model came out as the worst model for forecasting one-step ahead using both MAE and RMSE as the metrics for evaluation as shown in Figure 6. On the other hand, the ARIMA model proved to have the least error among all the other developed models. The GRU model also performed relatively better than the RNN model and the LSTM model as shown in Figure 6.

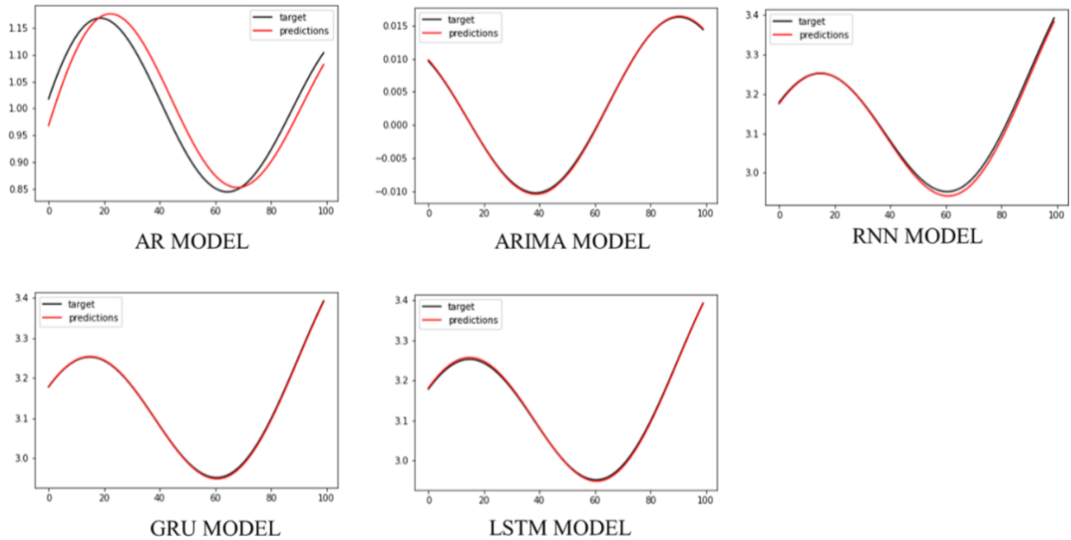

Figure 5: One-step forecasting of the various models

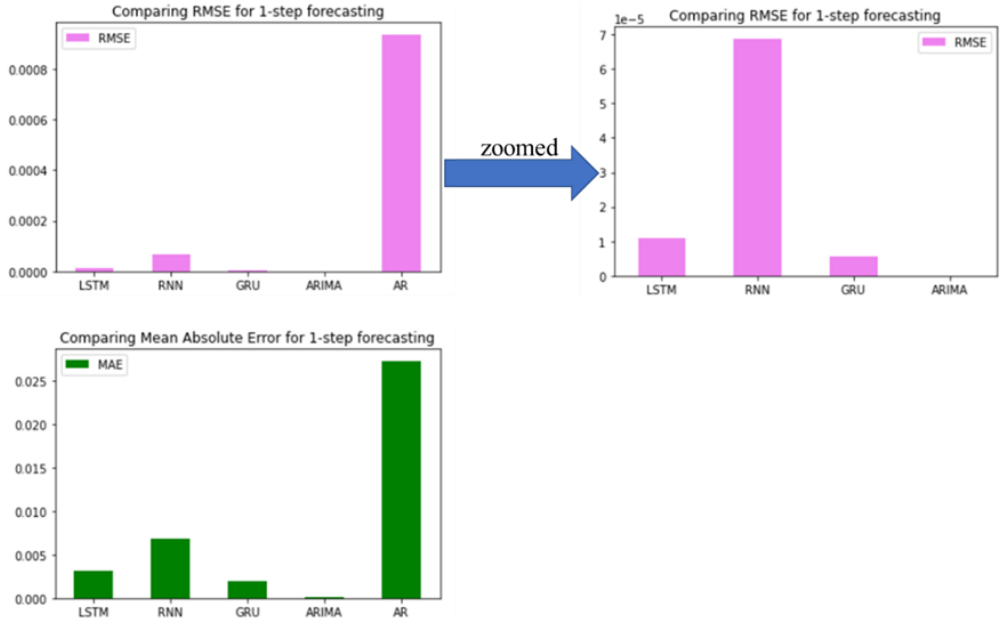

Figure 6: Evaluating the performance of the developed models in forecasting one-time step ahead.

### *4.4 Comparing 5-steps forecasting for all developed models:*

Figure 7 shows the various graphs for five-step forecasting for the developed models. The effectiveness of each model was evaluated using MAE and RMSE. The AR model came out as the worst model for forecasting five-steps ahead using both MAE and RMSE as the metrics for evaluation as shown in Figure 8. On the other hand, the GRU model proved to have the least error (see Figure 8) among all the other developed models. The RNN model also performed relatively better than the ARIMA model and the LSTM model as shown in Figure 8.

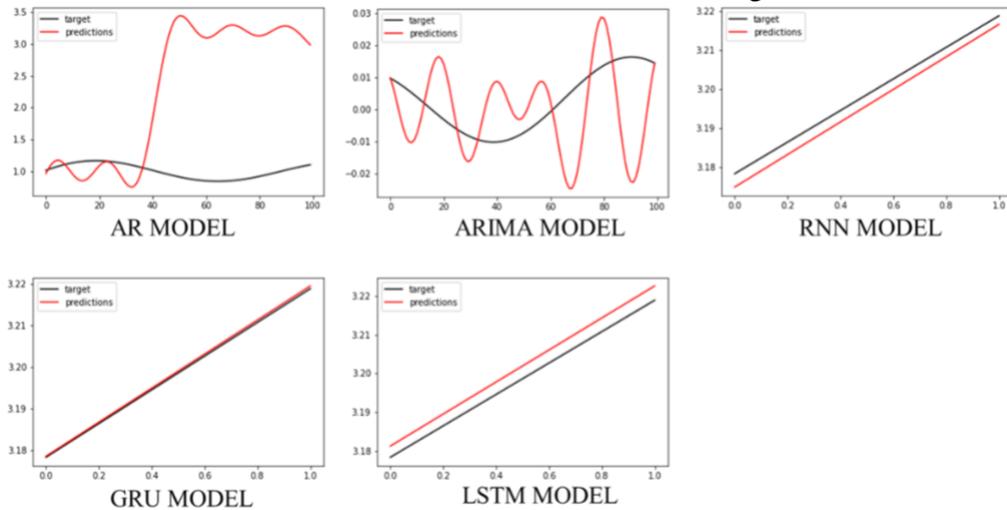

Figure 7: five-steps forecasting of the various models

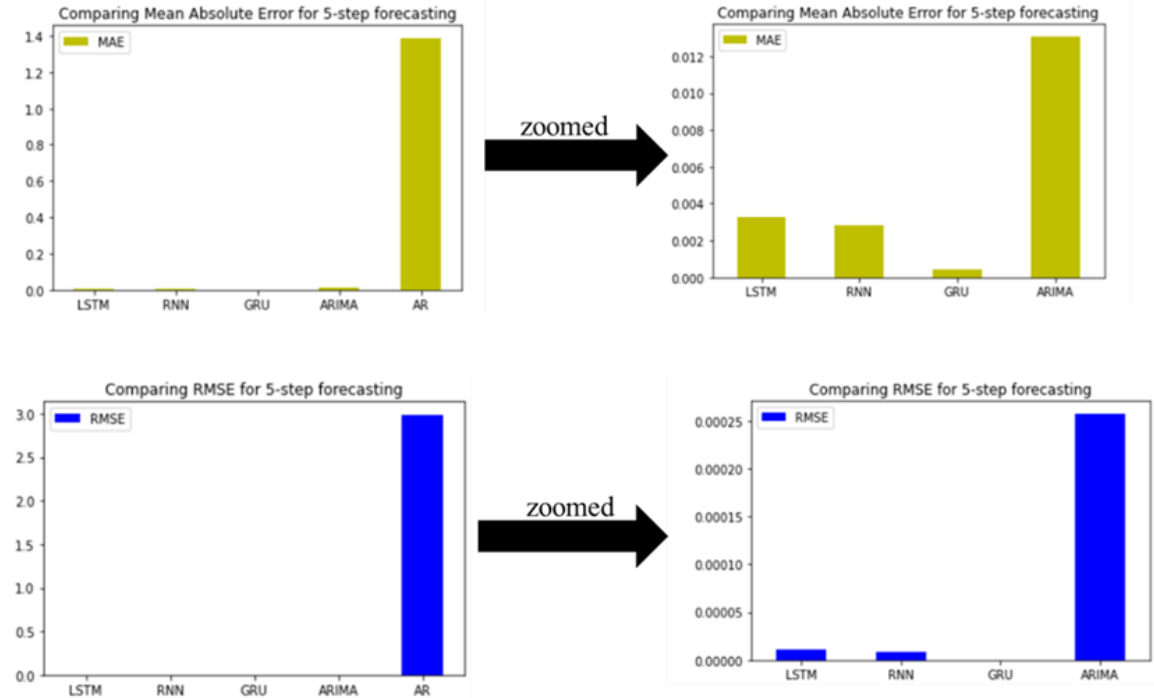
Figure 8: Evaluating the performance of the developed models in forecasting five-time steps ahead.

## 5. Conclusion
Some major findings from this study include.
1. The number of RNN layers affected the time taken to build the model. An increase in the number of RNN layers translate to a longer time to build the model. It also increases the probability of overfitting in the model
2. Increasing the number of hidden layers improved the model accuracy
3. The learning rate also influenced the time taken for the model to build. The learning rate was varied from 0.0001-0.01, the smaller the learning rate, the longer it took the model to build.

In conclusion, the ARIMA model proved to be the best model architecture for travel time prediction and forecasting.

**References**

[1] Idé, T. and Kato, S. "Travel-Time prediction using Gaussian process regression: a trajectory-based approach". SIAM Intl. Conf. Data Mining (2009).
[2] Wu, C-H., Wei, C-C., Ming-Hua Chang, M-H., Su, D-C. and Ho, J-M. "Travel Time Prediction with Support Vector Regression". Proc. Of IEEE Intelligent Transportation Conference. October 2003 pg. 1438-1442.
[3] Daniel Billings, Student Member, IEEE and Jiann-Shiou Yangt, Senior Member, IEEE, "Application of the ARIMA Models to Urban Roadway Travel Time Prediction - A Case Study", 2006 IEEE International Conference on Systems, Man and Cybernetics, 2006, Volume: 3, pp. 2529 - 2534

[4] GAO, Jianming HU*, Hao Zhou and Yi Zhang, "Travel Time Prediction with Immune Genetic Algorithm and Support Vector Regression", 2016 12th World Congress on Intelligent Control and Automation (WCICA), 2016, pp. 987 – 992

[5] Yanjie Duan1;2;3, Yisheng Lv1 and Fei-Yue Wang1, "Travel Time Prediction with LSTM Neural Network", 2016 IEEE 19th International Conference on Intelligent Transportation Systems (ITSC), 2016, pp. 1053 - 1058

[6] Yangdong Liu, Yizhe Wang, Xiaoguang Yang *, Linan Zhang, "Short-term travel time prediction by deep learning: A comparison of Different LSTM-DNN Models", 2017 IEEE 20th International Conference on Intelligent Transportation Systems (ITSC), 2017, pp.1 - 8

[7] Spaccapietra, S., Parent, C., Damiani, M. L., de Macêdo, J.A., Porto, F., and Vangenot, C. A conceptual view on trajectories. Data & Knowledge Engineering 65, 1 (2008),126-146. Including Special Section: Privacy Aspects of Data Mining Workshop (2006) - Five invited and extended papers.

[8]. Kwakye, K., Seong, Y., & Yi, S. (2020, August). An Android-based mobile paratransit application for vulnerable road users. In *Proceedings of the 24th Symposium on International Database Engineering & Applications* (pp. 1-5).